\documentclass{article}
\usepackage{ijcai24}

\usepackage{times}
\usepackage{soul}
\usepackage{url}
\usepackage[hidelinks]{hyperref}
\usepackage[utf8]{inputenc}
\usepackage[small]{caption}
\usepackage[dvipdfmx]{graphicx}
\usepackage{amsmath}
\usepackage{amsfonts}
\usepackage{amssymb}
\usepackage{amsthm}
\usepackage{booktabs}
\usepackage{algorithm}
\usepackage{algorithmic}
\usepackage[switch]{lineno}

\usepackage{subfigure}
\usepackage{tabularx}
\usepackage{multirow}
\usepackage{color}

\title{Action is the primary key: a categorical framework for episodic memories and logical reasoning}

\author{
    Yoshiki Fukada
    \affiliations
    Toyota Motor Corporation
    \emails
    yoshiki\_fukada@mail.toyota.co.jp
}

\begin{document}

\maketitle

\begin{abstract}
This study presents data format of episodic memory for artificial intelligence and cognitive science. 
The data format, named cognitive-logs, enables rigour and flexible logical reasoning.  
Cognitive-logs consist of a set of relational and graph databases. 
Cognitive-logs store an episodic memory as a graphical network 
that consist of ``\textit{actions}" represented by verbs in natural languages 
and ``\textit{participants}'' who perform the actions. 
These objects are connected by arrows (morphisms) that bind 
each action to its participant and bind causes and effects. 
The design principle of cognitive-logs refers cognitive sciences especially in cognitive linguistics. 
Logical reasoning is the processes of comparing causal chains in episodic memories with known rules 
which are also recorded in the cognitive-logs. 
Operations based on category theory enable such comparisons between episodic memories or scenarios. 
These operations represent various inferences 
including planning, comprehensions, and hierarchical abstractions of stories. 
The goal of this study is to develop a database-driven artificial intelligence 
that thinks like a human 
but possesses the accuracy and rigour of a machine. 
The vast capacities of databases (up to petabyte scales in current technologies) enable 
the artificial intelligence to store a greater volume of knowledge 
than neural-network based artificial intelligences. 
Cognitive-logs also serve as a model of human cognition mind activities. 

\end{abstract}

\section{Introduction}

%hoge\cite{dummy}
Logical reasoning on computers has been an important research topic. 
%in the field of computer science. 
%Reasoning processes like ``\textsf{if A then B}" 
%have been computerized for a long time. 
%A notable achievement in this area is the development of 
The logic programming language Prolog\cite{Kowalski1988} 
and the application into expert systems\cite{Feigenbaum1984} 
were notable achievements. 
%The Prolog language has been successfully applied to various systems, 
%such as expert systems\cite{Feigenbaum1984}. 
%
Representations of knowledge in graphical network has also been a long-standing area. 
%of study. 
Semantic networks were introduced\cite{Woods1975}. 
Later, a more restricted concept---knowledge graphs---was defined\cite{Bakker1987}. 
By applying category theory, ontology logs, or ologs, were introduced\cite{Spivak2012}. 
An application of category theory into cognitive science is also presented\cite{Fuyama2020}. 
%
%While logic programming has a long research history and numerous successful outcomes, 
Despite its long history, 
application of logic programming into real-world scenarios has been limited. 
%
%In a logic depicted by the ``\textsf{if A then B}" formula, 
%\textsf{A} and \textsf{B} are not merely objects but scenarios, 
%necessitating comparisons of real episodic memories to these scenarios. 

The developments in neural network-based artificial intelligence show significant progress. 
An ambitious project\cite{Yamakawa_2021} seeks to reproduce a brain. 
%as an assembly of neural-network based machine-learning. 
The invention of the transformer\cite{vaswani2017attention} 
and developments of Large Language Models (LLMs) are a prominent milestone. 
However, as a vision for new artificial intelligence pointed out\cite{LeCun2022}, 
current LLMs do not yet seem to replicate human thinking. 
%Services that use large-scale language models (LLMs) 
%provide ``wonderful" responses, giving the impression that 
%``logical reasoning in artificial intelligence has been realized." 
The learning processes of LLMs require an enormous amount of data. 
In contrast, humans are capable of learning from even a single experience. 
As several studies had shown, accuracy of logical reasoning in LLMs 
are quite limited\cite{Mizadeh2024} and \cite{Wang2024}.
%accuracy of reasoning in these LLMs is also questioned. 
%Several studies show that accuracy of their logical reasoning 
%are quite limited\cite{Mizadeh2024,Wang2024}. 
Moreover, these LLMs are ``black boxes," 
hence lack explainability. 

It is widely believed that our brains are Bayesian, 
however, arguments also exist\cite{Bain2016}. 
If we have inherent ``world-models", 
it limits the degree of freedoms of our learning machines 
%constraints from it significantly reduce the degree of freedoms of our learning machines 
and allow adaptation with limited amount of learning data. 
%Animals including us show inherent abilities. 
%Some arthropods exhibit amazing intelligences. 
%Even common spiders demonstrate impressive abilities in constructing their webs. 
An experimental study unveiled that 
human infant can count numbers\cite{Wynn1992}. 
%However, 
This experiment suggests that we have an inherent world-model 
that ``numbers of objects are preserved" rather than ability of counting. 
%It is remarkable that a vision for world-model-based intelligent machines 
%was presented by one of the AI-legend\cite{LeCun2022}. 

Common spiders demonstrate impressive abilities in constructing their webs. 
Human have extremely large brains, 
%Human brains have extremely large frontal lobes, 
with the number of neurons being $10^6$ times greater than that of spiders. 
%arthropods. 
%What accounts for the difference in brain size? 
Perhaps, this brain size is for logical reasoning that other creatures do not poses. 
However, our reasoning is still awkward; 
we can't calculate well even for only a few digits of numbers. 
This inconvenient truth tells us that 
neural-network architectures are not efficient for logical reasoning. 
%Logical reasoning is something these small creatures do not poses. 
%Perhaps neural-network architecture is not efficient for logical reasoning. 

%In the construction of the new framework, 
This research exploits two distinct academic fields. 
Category theory\cite{spivak_text} is a fascinating mathematical theory. %for graphical networks. 
A conversion of a graphical network into a category allows 
abstract handling and precise analysis. 
Natural languages seem to be strongly related to our logical reasoning 
and cognition mechanisms. 
Cognitive linguistics\cite{Lakoff1987} 
seeks to understand human cognition through linguistics. 
%Their most significant finding is the 
Common of grammatical features across different languages 
%Even languages of isolated ethnicities have common features with other languages. 
%This 
indicate common cognitive mechanisms among human 
and should be attributed to inherent world-models 
rather than being the result of learnings.  

According to the discussions above, 
this study determined the design philosophy of the framework as follows: 
%---cognitive-logs---as follows: 
%
\begin{itemize}
 \item Not to use neural-networks. 
 \item Uses graphical network. 
 \item Learning is NOT statistical. 
% \item Applies category theory
% \item Uses knowledge of cognitive sciences especially cognitive linguistics\cite{Lakoff1987}
\end{itemize} 
%
%Thus, cognitive-logs were designed aiming that 
%the structures of cognitive-logs represent structures of the world in our world-model. 
%
%The following sections illustrates 
%the construction of cognitive-logs and its operation with some reasoning examples. 
%
The framework, named cognitive-logs, consist of e-logs, be-logs, and s-logs that 
record episodic memories, static relationships between objects, 
and knowledge of laws and rules, respectively. 
%
%It will be shown that 
Various inference processes---%
abstractions, deductions, inductions, and analogies---%
%, and metaphors---%
can be modelled using cognitive-logs and operations on them. 
The details are described below. 

%%%%%%%%%%%%%%%%%%%%%%%%%%%%%%%%%%%%%%%%%%%%%%%%%%%%%%%%

\section{Categorical description of episodic memory}

\subsection{Graphical image and it's problems}

Consider a simple event, ``Bob loves Alice." 
If one draws a graphical network representing this situation, 
the network may look as follows: 
\begin{equation}\label{bobloves}%1
\begin{array}{ccc}
 & \text{loves} & \\
 \ulcorner\text{Bob}\urcorner & \longrightarrow & \ulcorner\text{Alice}\urcorner.
\end{array}
\end{equation}
This graphical structure may represent the typical subject-verb-object (SVO) sentence structure. 
However, this structure has problems. 
First, an action does not always have an object. 
%subject-verb-object components. For example, 
An action represented by an intransitive verb has only subject and verb. 
Second, this structure is not symmetric with respect to Bob and Alice. 
This asymmetry makes the system complex. 
Third, representation of causal relationships is unclear. 
Fourth, this network does not describe ``\textit{love.}" 
Thus, the subject-to-object graphical structure such as (\ref{bobloves}) 
is not universally applicable.

\subsection{Prerequisite assumptions}

%The motivation of this study is to recreate the thought process of humans. 
Human cognitions are classified into things-like elements and process-like elements, 
which correspond to nouns and verbs, respectively\cite{Langacker1987}. 
%Their boundaries between nouns and verbs are vague\cite{Ross1973}. 
%In most natural languages, verbs can be converted into nouns. 
%A verb represents an action, and the action represents an elemental event. 
%Although actions do not exist as real objects, 
%our brains perceive actions as entities. 
%Perhaps, our brains have ``action cells" similar to the grandmother cells. 
%In Japanese spoken language, subjects are often omitted, but verbs are not. 
As \cite{tomasello1999} pointed out that, 
actions are the primary entities in an event.

Corresponding to nouns and verbs, 
the primary component of an event is the action and the participant. 
Here, ``participants" is the term in cognitive linguistics for subjects and objects. 
The primary relationships between these components are 
connection between actions and participants (who performs the action) 
and causations between events. 
Since an action represents the smallest unit of of an event, 
the causations must be placed between actions. 

Based on consideration above, 
following rules concerning action and participants were assumed: 
\begin{itemize}
 \item An action has a participant \textbf{who} performs the action. 
 \item An action has an action that is the \textbf{cause} of it. 
 \item An action has an action that is the \textbf{effect} of it. 
\end{itemize}
It should be noted that these rules are parts of our world-model, 
and there is a remarkable rule behind that ``actions are discrete." 
This discretization of actions allows the categorical operations described below 
and enables logical reasoning.

\subsection{``\textit{do}"--``\textit{be done}" decomposition}

\begin{figure}
%\centering 
  \includegraphics[width=8cm]{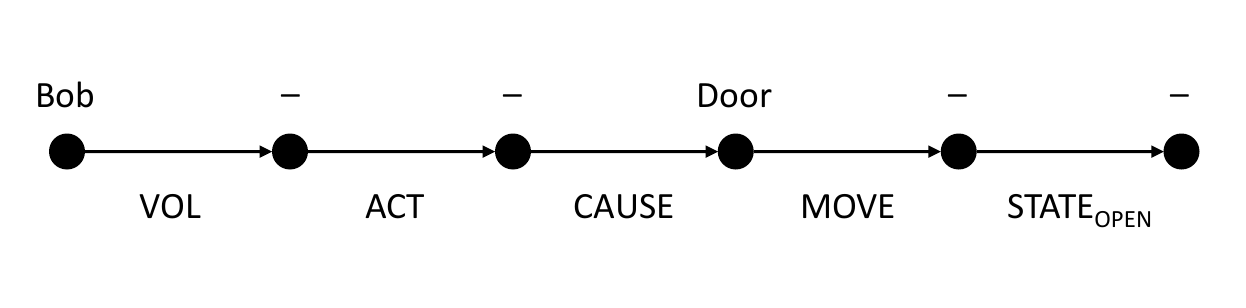}
 \caption{Causal chain model of ``Bob opened the door" in cognitive linguistics. 
             The symbols ``VOL" and ``ACT" are abbreviations of ``volition" and ``action," respectively. 
             VOL represents existence of Bob's intention to open the door. 
             A causal chain model corresponds well to an e-log.  
      }
  \label{schema}
\end{figure}

Figure \ref{schema} shows a preceding concept from cognitive linguistics: 
causal chain model\cite{croft1991,Ohori}. 
The event ``Bob opened the door." is decomposed into 
two elemental events: ``Bob let the door open" and ``The door opened."

Alice is in a situation of ``\textit{being loved}" 
because Bob loves Alice. 
This consideration leads the idea ``\textit{do}"--``\textit{be done}" decomposition. 
Namely, the action ``\textit{loves}" is decomposed 
into the pair ``\textit{loves}" and ``\textit{is loved.}" 
This decomposition enables 
a symmetric description concerning Bob and Alice 
and fulfilling the requirement that an action has a participant.

\subsection{Episode-logs or e-logs}

\begin{figure}%2
\centering 
  \includegraphics[width=7cm]{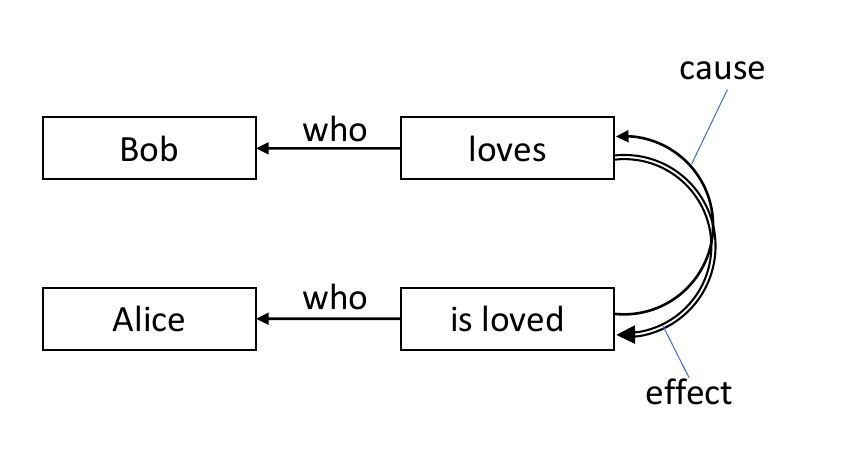}
 \caption{E-log depicting ``Bob loves Alice" 
%             in the form of a category of elements. 
%             This network satisfies the conditions of a category. 
%             This figure omits indications of ``\textit{nobody,}'', ``\textit{unknown,}'' 
 %             ``\textit{nothing,}'' and identity morphisms. 
          %   Throughout this paper, 
          %   an individual object is indicated with a rectangular text-box 
          %   or corner-symbols $\ulcorner\:\:\:\urcorner$ 
          %   in the form of a category of elements, 
          %   and an arrow of ``\textit{effect}" is indicated as a double line.  
      }
  \label{bobalice}
\end{figure}

Category theory is useful for graphical data structure. 
A category consists of objects and arrows (morphisms), 
where an arrow points from an object (domain) to an object (co-domain). 
The objects of the category that represents episodic memories are actions and participants.

The construction of e-log adopts the principal of ologs\cite{Spivak2012}. 
%Ologs are categories of database which the arrows are functions, 
%and convert relationships between conceptual objects into category. 
%
By applying ologs, the relationships between action and participants 
generate a category that is: 
\begin{itemize}
\item An action emanates an arrow of ``\textit{who}", an arrows of ``\textit{cause}", 
 and an arrows of ``\textit{effect}" 
 into a participant who perform the action, an action that is the cause of it, 
 and an action that is the effect of it, respectively. 
\end{itemize}
Since causes precede effects, 
arrows of ``\textit{cause}" point to the past, 
and arrows of ``\textit{effect}" point to the future. 
Hence, these two types of arrows are distinguished by referring temporal orders between actions. 

The e-log for the event of Bob and Alice is illustrated in Fig. \ref{bobalice}. 
Virtual actions 
``\textit{unknown}", ``\textit{nothing}", and ``\textit{nobody}", 
and corresponding arrows are added to accommodate the rules above. 
Any object in a category has its identity morphism (arrow). 
This paper only indicates identity morphisms when it is necessary to be shown.

\subsection{Causal relationships}

Linguistics studies on number of languages suggest that 
causality is an essential part of our cognition (\cite{tomasello1999} and \cite{hooperThompson1980}). 
It can be said that e-logs are mathematically rigorous causal chain model. 

An action (elemental event) may require multiple conditions to occur. 
An action is able to receive multiple ``\textit{effect}" arrows. 
Using this property, arrows of ``\textit{effect}" represent necessary conditions. 
On the other hands, an action emanates only one arrow of ``\textit{cause}." 
An arrow of ``\textit{cause}" points the ``primary" cause 
or ``the last piece" of the necessary condition of an action. 
A network of these ``\textit{cause}" and ``\textit{effect}" 
may depict a why-because-graph\cite{ladkin1998}.

\subsection{Trivial causal-relationships}

\begin{figure}%4
\centering
  \includegraphics[width=7.cm]{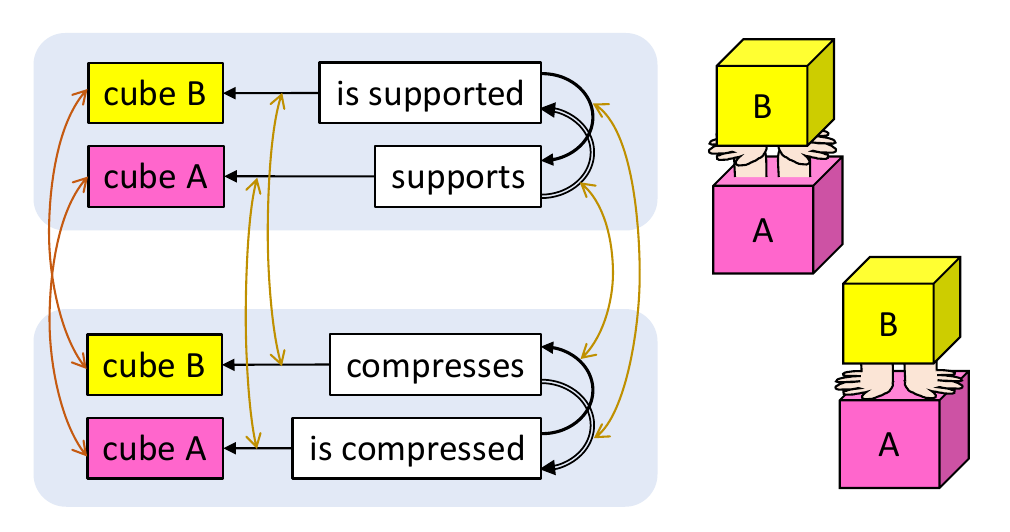}
 \caption{Two perspectives of a situation ``Cube B is on top of Cube A." 
             %(a) two distinct personifications of cubes without volition. 
             These e-logs are equivalent. 
             %(b) E-log with volition and its removal. 
 }
  \label{volition}
\end{figure}

The events ``Bob loves" and ``Alice is loved" are in a ``\textit{do}"--``\textit{be done}" relationship. 
This research regards such a causal relationship 
%based on ``\textit{do}"--``\textit{be done}" decomposition 
as a ``trivial causal relationship." 
In some trivial relationships, each ``\textit{do}" and ``\textit{be done}" occur simultaneously. 
It means that the ``\textit{cause}" and ``\textit{effect}" arrows can be exchanged.  
%This property allows the two events ``\textit{do}" and ``\textit{be done}" being equivalent, 
%and exchange the subject and object. 
%
Figure \ref{volition}a illustrates two perspectives of a situation where 
``Cube $B$ is on top of Cube $A$." 
%that are ``Cube $B$ compresses cube $A$", and ``cube $A$ supports cube $B$," respectively. 
%Since these ``\textit{cause}" and ``\textit{effect}" arrows are exchangeable. 
%The correspondence of objects and arrows (i.e., functor) shows their equivalence. 
These ``\textit{cause}" and ``\textit{effect}" arrows are exchangeable, 
and it allows to handle  
``Cube $B$ compresses cube $A$", and ``cube $A$ supports cube $B$,"
equivalently.

\subsection{Examples of categorical description}

\begin{figure}%3
\centering 
  \includegraphics[width=6cm]{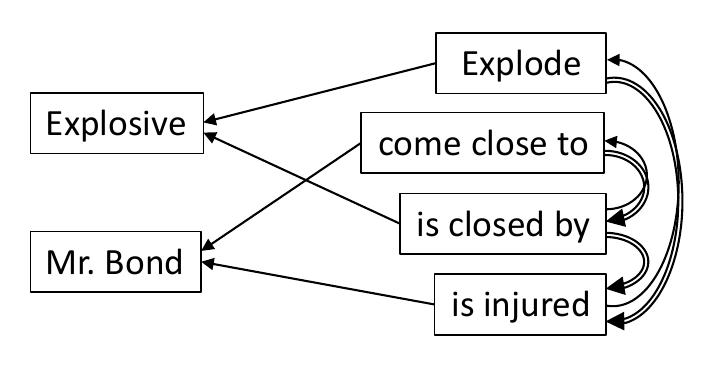}
 \caption{E-log for an explosion and injury. 
      }
  \label{explosionelog}
\end{figure}

Figure \ref{explosionelog} shows an event, 
Mr. Bond, was injured due to an explosion. 
The necessary conditions for Mr. Bond's injury were the explosion and his proximity to it. 
The explosion is the primary cause.

\subsection{E-logs as category of sets}

\begin{figure}%4
\centering 
  \includegraphics[width=8cm]{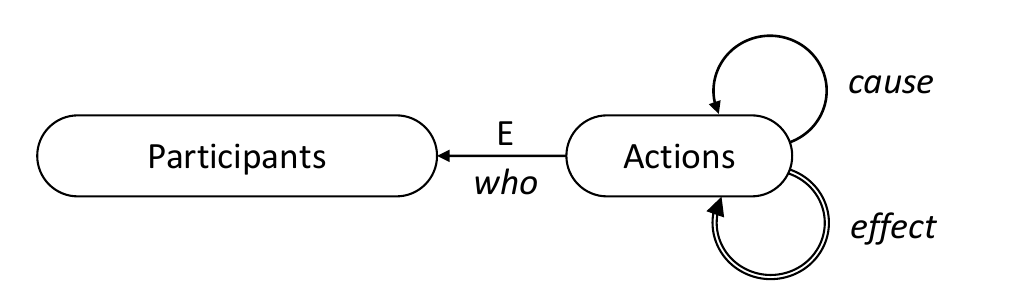}
 \caption{Structure of an e-log as a category of set. 
          %   Throughout this paper, a set of objects is indicated using a oval. 
      }
  \label{elog}
\end{figure}

An e-log can be converted into a category of sets. 
Bob and Alice are members of the set ``\textit{Participants,}" 
and ``\textit{nobody}" is as well. 
The actions ``\textit{loves,}" ``\textit{is loved,}" ``\textit{nothing}" and ``\textit{unknown}" 
are members of the set ``\textit{Actions.}" 
In realistic implementation, these objects are expressed as tokens. 
It should be noted that each action must be unique. 
The structure of e-logs as a category of sets is illustrated in Fig. \ref{elog}. 

E-log as a category of sets consist of these two sets above and following functions:
\begin{equation}\label{elog_category_who}%
 \mathit{who} : \mathit{Actions} \to \mathit{Participants}.
\end{equation}%
\begin{equation}\label{elog_category_cause}%
 \mathit{cause} : \mathit{Actions} \to \mathit{Actions}.
\end{equation}%
\begin{equation}\label{elog_category_effect}%
 \mathit{effect} : \mathit{Actions} \to \mathit{Actions}.
\end{equation}

\subsection{E-logs as databases}

A relational database\cite{Codd1970} satisfies the requirements of a category. 
A category of sets can be converted into a relational database\cite{databaseiscategory} and \cite{databaseiscategory2}. 
A database consists of tables. 
A domain (object that emanates arrows) of the category is a primary key in a table, 
and its codomains (object where arrows points) are alternate keys of the table. 
An e-log is converted into a relational database 
which the primary key is ``\textit{actions}". 
%and the foreign keys of ``\textit{who,}" ``\textit{cause,}" and ``\textit{effect.}" 
The relational table for the ``Bob loves Alice" event is shown in table \ref{BobAlicetable}. 
%The indications in the table using English words are for the clarity of readers. 
In a realistic implementation, entities are recorded using unique IDs. 

With the spread of social networking services, 
current databases technology is now able to handle 
enormous amount of data, even on a petabyte scale\cite{TAO}. 
This scale of data capacity dwarfs that of any other neural-network-based artificial intelligence. 
The potential of cognitive-logs is promising. 

%%%%%%%%%%%%%%%%%%%%%%%%%%%%%%%%%%%%%%%%%%%%%%%%%%%%%%%%%%%%%%

\section{Similarity, association, classification, and description of characteristics}

\subsection{Be-logs}

\begin{table}
 \caption{Database of ``Bob loves Alice"}
 \begin{center}
  \begin{tabular}{c||c|c|c} \hline
     Action   & Who      & Cause     & Effect   \\  \hline
     unknown & nobody & unknown  & unknown \\
     loves     & Bob      & unknown  & is loved  \\
     is loved  & Alice     & loves      & nothing  \\ 
     nothing  & nobody  & nothing    & nothing  \\ \hline
  \end{tabular}\label{BobAlicetable}
 \end{center}
\end{table}

Our basic cognition involves classification, such as ``a pigeon is a bird." 
Such classifications are based on similarity recognition, 
which is implemented through associations. 
These cognitions are mostly expressed using a \textit{be}-verb 
and hold a special (more essential) position. 
Be-logs are categorical networks that describe static relationships such as  
similarity, association, classification, and description of characteristics. 
The name ``be" represents \textit{be}-verbs. 

Similarity recognition of human has the following characteristics.

\begin{itemize}
\item Asymmetry: The similarity of $A$ to $B$ 
 is not necessarily equal to the similarity of $B$ to $A$\cite{Tversky1977}.  
\item Composition is not guaranteed: $A$ being similar to $B$, and $B$ being similar to $C$, 
 does not guarantee that $A$ is similar to $C$.
\end{itemize}

%For the description of similarities and classifications, 
Since ``\textit{do}"--``\textit{be done}" decomposition is redundant for this cognition, 
%because these relationships are static. 
%For such relationships, 
a graphical image between objects is appropriate. 
Applying the theory of category of graph\cite{spivak_text}, 
the category of the relationship is illustrated as follows: 
\begin{equation}\label{resemble}%2
 \begin{array}{cccccc}
  &source&&target& \\
   \ulcorner\text{A}\urcorner & \longleftarrow & \ulcorner\text{resembles}\urcorner &
   \longrightarrow & \ulcorner\text{B}\urcorner \text{.}\\
\end{array}
\end{equation}

Other relationships, such as ``An apple is a fruit,'' can be represented in the same way: 
\begin{equation}\label{isidentifiy}%3
 \begin{array}{cccccc}
  &source&&target& \\
   \ulcorner\text{Apple}\urcorner & \longleftarrow & \ulcorner\text{is (identified as)}\urcorner &
   \longrightarrow & \ulcorner\text{fruit}\urcorner \text{.}\\
\end{array}
\end{equation} 

Descriptions of characteristics are also depicted using the \textit{be}-verb. 
For example, the description ``The apple is red" is represented using this form: 
\begin{equation}\label{is}%4
 \begin{array}{cccccc}
  &source&&target& \\
   \ulcorner\text{Apple}\urcorner & \longleftarrow & \ulcorner\text{is}\urcorner &
   \longrightarrow & \ulcorner\text{red}\urcorner \text{.}\\
\end{array}
\end{equation}

\begin{figure}%6
\centering 
  \includegraphics[width=7cm]{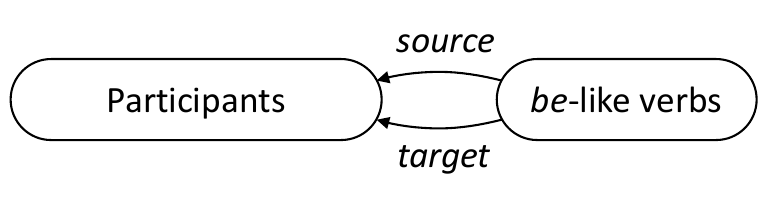}
 \caption{Be-log denotes classifications, similarities, or associations between objects. 
      }
  \label{belog}
\end{figure}

The following verbs or phrases---``be," ``resemble," ``is similar to," ``evoke," ``be reminiscent of," 
and many others---% 
express cognitions of similarities, classifications, descriptions of characteristics, or associations. 
This research regards these verbs and phrases as "\textit{be}-like verbs." 
Be-log records these relationships between objects. 
We recognize similarities not only between nouns but between \textit{Actions}. 
Similarities between actions, 
or features of an action (``strongly''/``softly'', ``fast''/``slowly'', and others) 
are also stored in be-logs. 
Figure \ref{belog} illustrates the structure of be-logs.

Be-log as a category of sets consist of following functions:
\begin{equation}\label{belog_category_source}%
 \mathit{source}: \mathit{be}\text{-}\mathit{like}\text{-}\mathit{verbs} \to \mathit{Participants}.
\end{equation}
\begin{equation}\label{belog_category_target}%
 \mathit{target}: \mathit{be}\text{-}\mathit{like}\text{-}\mathit{verbs} \to \mathit{Participants}.
\end{equation}
Here, $\mathit{be}\text{-}\mathit{like}\text{-}\mathit{verbs}$ is a set of \textit{be}-like verbs. 
It should be noted that each \textit{be}-like verb is unique and also expressed as tokens. 
There must be type information of each \textit{be}-like verb 
and details of similarities, identifications or other relationships of two participants. 
Attribution of these information as category or database is future consideration.

\begin{figure}%7
\centering
% \subfigure[Future prediction]{
  \includegraphics[width=5cm]{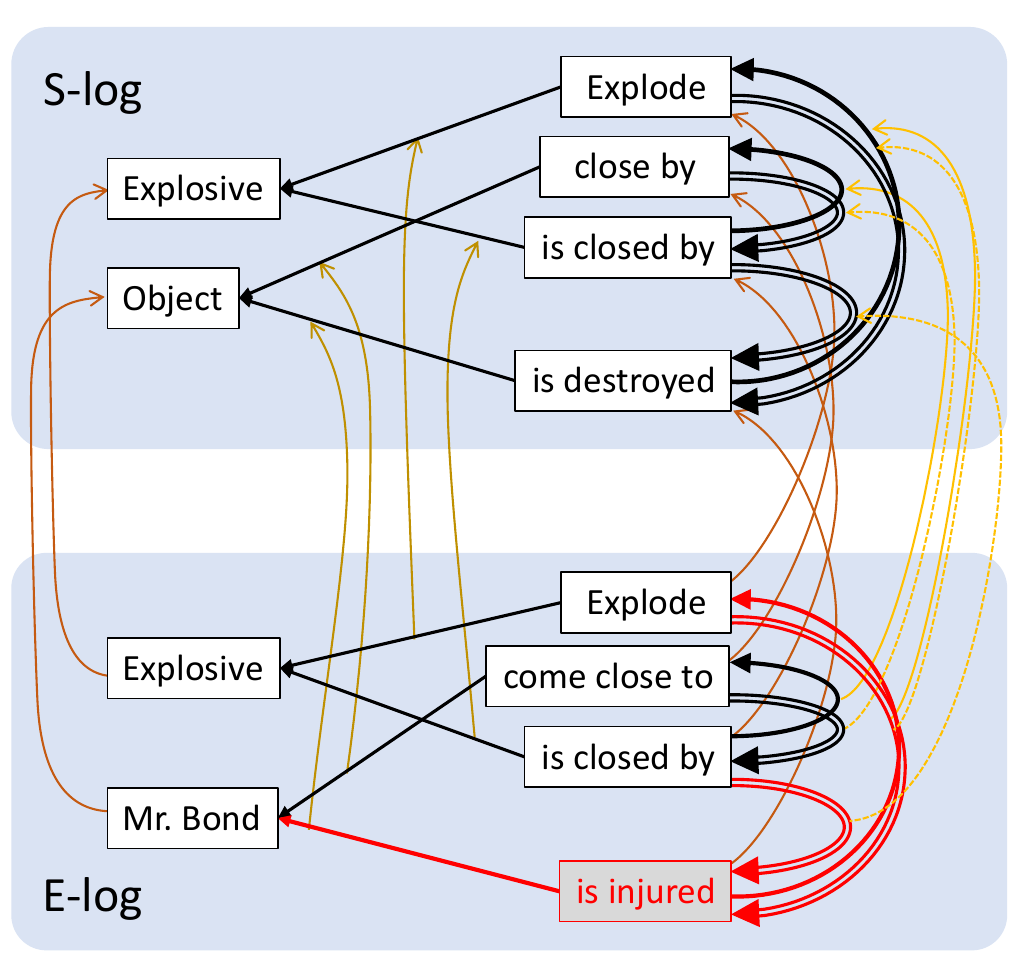}%}% 
 \caption{Functors concerning a Mr. Bond's event. 
%injury according to the law 
%             that ``an explosion destroys things nearby" 
%             
             Prediction of Mr. Bond's injury as filling in incomplete parts of the functor. 
                %from the e-log to the s-log
                %predicts occurrences in the near future. 
             %b,c) Abstractions of event into different scenarios. 
%             a) the functor from the e-log into the s-log omits the tower and the related actions. 
%                The actions ``\textit{destroy}" and ``\textit{is destroyed}" are not in the e-log. 
%             b) the actions ``\textit{destroy}" and ``\textit{is destroyed}" 
%                are copied to the e-log as future items. 
%                Then the functor is complete.  
    }
  \label{explosionfunctor}
\end{figure}

\subsection{Types of \textit{be}-like verbs}

\begin{table*}
 \caption{Types of \textit{be}-verbs}
 \begin{center}
  \begin{tabular}{c||c|c|c} \hline
     Type of & Description      & Representation & Example  \\  
     \textit{be}-verb &           & in English         &  in English  \\ \hline
     Be1  & Identification        & ``be"     & He is Bob.     \\
     Be2  & Equivalence          & ``be"     & Bob is (the same with) Mike.     \\ 
     Be3  & Classification        & ``be"     & Bob is a human.      \\ 
     Be4  & Characteristic       & ``be"     & The Apple is red.      \\ 
            &                           & ``has characteristic" & The apple has red color. \\
            &                           & ``has component"    & A bird has a beak. \\
     Belong  & Belonging          & ``belong to"     & Daddy's car.      \\ 
     Similar  & Similarity          & ``be similar to" & Mango is similar to apple. \\ 
                &                       & ``resemble"      & Mango resembles apple. \\
     Association  & Association & ``evoke"          & Mango evokes apple. \\ 
               &                             & ``be reminiscent of" & Mango is reminiscent of apple. \\ \hline
  \end{tabular}\label{typeofbeverbs}
 \end{center}
\end{table*}

%This research regards "\textit{be}-like verbs" as verbs 
%that do not indicate an event but a static relationship. 
It is interesting that the verb ``\textit{be}" has a wide range of uses. 
It should be noted that \textit{be}-verbs have another meaning of ``existence". 
Description of an existence of an object or a concept does not belong to a be-log, but belong to an e-log.   
Table \ref{typeofbeverbs} shows the types of \textit{be}-like verbs. 
Verbs indicating belongings and associations are static, 
hence they are regarded as \textit{be}-like verbs. 

\paragraph{Belongings}

In the observations of a child's language development\cite{Tomasello2000}, 
the use of possessives such as ``\textit{my}" or ``\textit{Daddy's}" developed at a quite early stage. 
%before developing subject--verb--object syntax. 
Therefore, this cognition of belonging is NOT ``Daddy owns a car," 
but ``The car is Daddy's." 
The idiom ``\textit{belong to}" seems to be a kind of \textit{be}-verb. 

\paragraph{Associations}

Association seems to be one of the elemental processes in our brains. 
When we see a mango, one may associate with an apple: 
%$\ulcorner\text{Mango}\urcorner  \xleftarrow{source} \ulcorner\text{evokes}\urcorner 
%   \xrightarrow{target}  \ulcorner\text{Apple}\urcorner$. 
%
\begin{equation}
 \begin{array}{cccccc}
  &source&&target& \\
   \ulcorner\text{Mango}\urcorner & \longleftarrow & \ulcorner\text{evokes}\urcorner &
   \longrightarrow & \ulcorner\text{Apple}\urcorner \text{.}\\
\end{array}
\end{equation} 

These \textit{be}-like verbs also present vector images. 
Such vector images seem universal in our cognition.

%%%%%%%%%%%%%%%%%%%%%%%%%%%%%%%%%%%%%%%%%%%%%%%%%%%%%%%%%%%%%%%%%%%%%%%%

\section{Functor between cognitive-logs}

Similarity recognition is one of our essential cognitive abilities. 
Similarity between events is also essential, and it enables logical reasoning. 
Abstractions of narratives---%
converting a narrative into a simpler form---% 
are also important cognition processes. 
Functors are powerful tools in category theory. 
A functor---a structure preserving map---between cognitive-logs represents these inference processes. 

%%%%%%%
%Consider e-logs $\mathcal{E}$ and $\mathcal{E'}$ as categories 
%and a functor $F$ from $\mathcal{E}$ to $\mathcal{E'}$. 
%The object function of $F$ is defined as follow:

%$Ob(F): Ob(\mathcal{E}) \to Ob(\mathcal{E'})$ 

%And a morphism function for objects $e_1,e_2 \in \mathcal{E}$ is defined as follow: 

%$Hom_F(e_1, e_2): Hom_\mathcal{E}(e_1,e_2) \to Hom_\mathcal{E'}(e_1', e_2')$. 

%where, $e_1'=F(e_1)$ and $e_2'=F(e_2)$, respectively. 
%%%%%%%

%Figure \ref{functor} illustrates the mapping above. 
%A functor is a structure preserving map. 
%Its exposition can be found in many excellent textbooks. 

\subsection{S-logs or scenario-logs as references}

One of the simplest logical inferences might be $1+1=2$. 
However, even this simple formula is invalid in Boolean algebra. 
This simple example shows that ``absolute logic" does not exist; 
instead, we ``refer to rules." 

We perform various complex logical inferences. 
%future prediction based on physical laws, 
%judgement applying ethical rules, 
%and constructing a new idea by joining multiple laws and rules. 
All of such inferences are based on ``comparison with reference scenarios 
that represent laws and rules." 
%In a broader sense, laws and rules are ``reference scenarios." 
%
%
%hence, the comparison between them is a functor.  
%Completeness of a functor represents the similarity 
%
%
%Considering the above, 
%s-logs or scenario-logs are presented as categories that have the same structure as e-logs. 
%This definition is not strict. 
%Just as we refer to our own experiences, 
%
%The details of s-logs will be discussed later. 
S-logs are variant of e-logs that represent such reference scenarios.

\subsection{Inferences according to laws}\label{inferences}

When Mr. Bond was %looking at the tower and 
walking around the site, 
an explosion suddenly occurred close to him... 
If one reads such a novel, he/she will surely predict injuries to Mr. Bond. 
Figure \ref{explosionfunctor} illustrates such a process. 
The s-log depicts that an explosion destroys things nearby. 
%Considering a functor from the s-log to the e-log, 
%unessential items---concerning the tower---are omitted, and 
In the e-log, the s-log's items ``\textit{destroy,}" and ``\textit{is destroyed}" are not in the e-log. 
Since there are causal relationships, 
these items are ``things that happen in the future." 
Thus, future events are found as missing items in the e-log.

\subsection{Abstraction}

\begin{figure}
\centering
 \subfigure[Abstraction into ``Enemy attack Mr. Bond"]{
  \includegraphics[width=4cm]{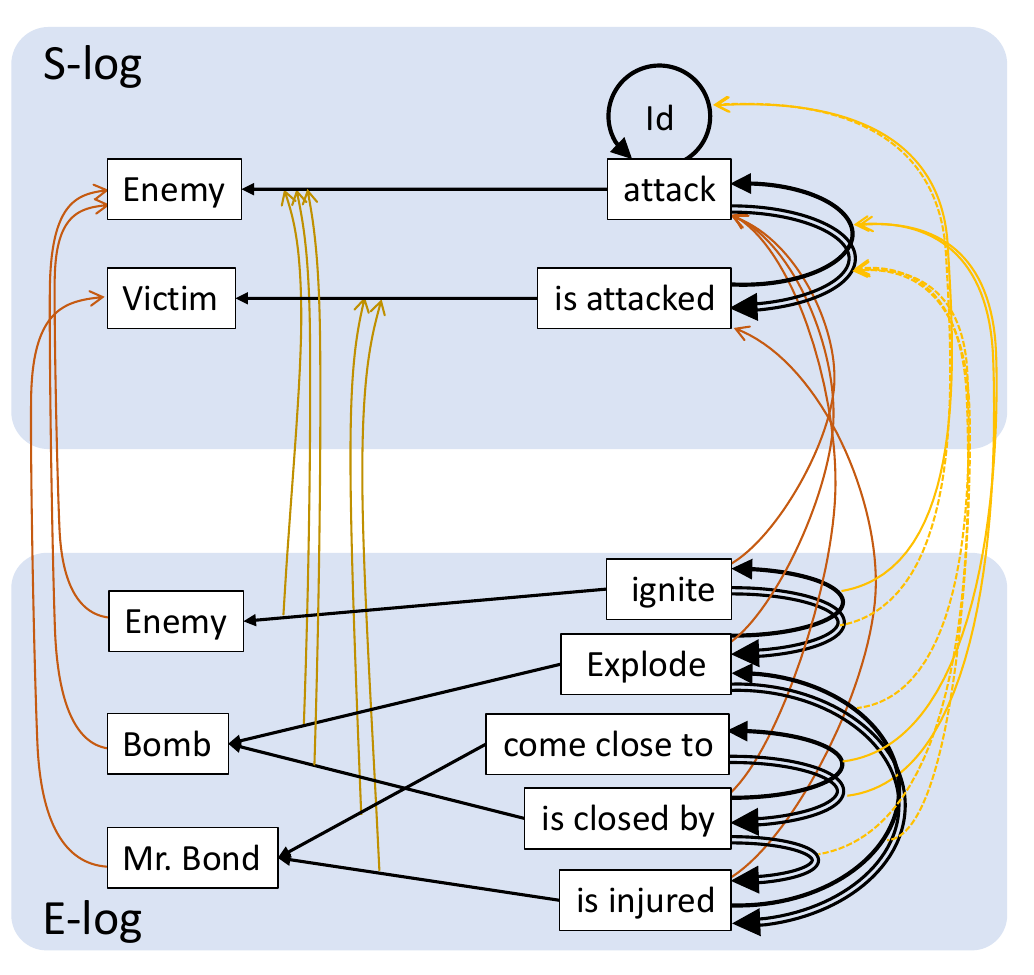}}% 
 \subfigure[Abstraction into ``Enemy bombard the site"]{
  \includegraphics[width=4cm]{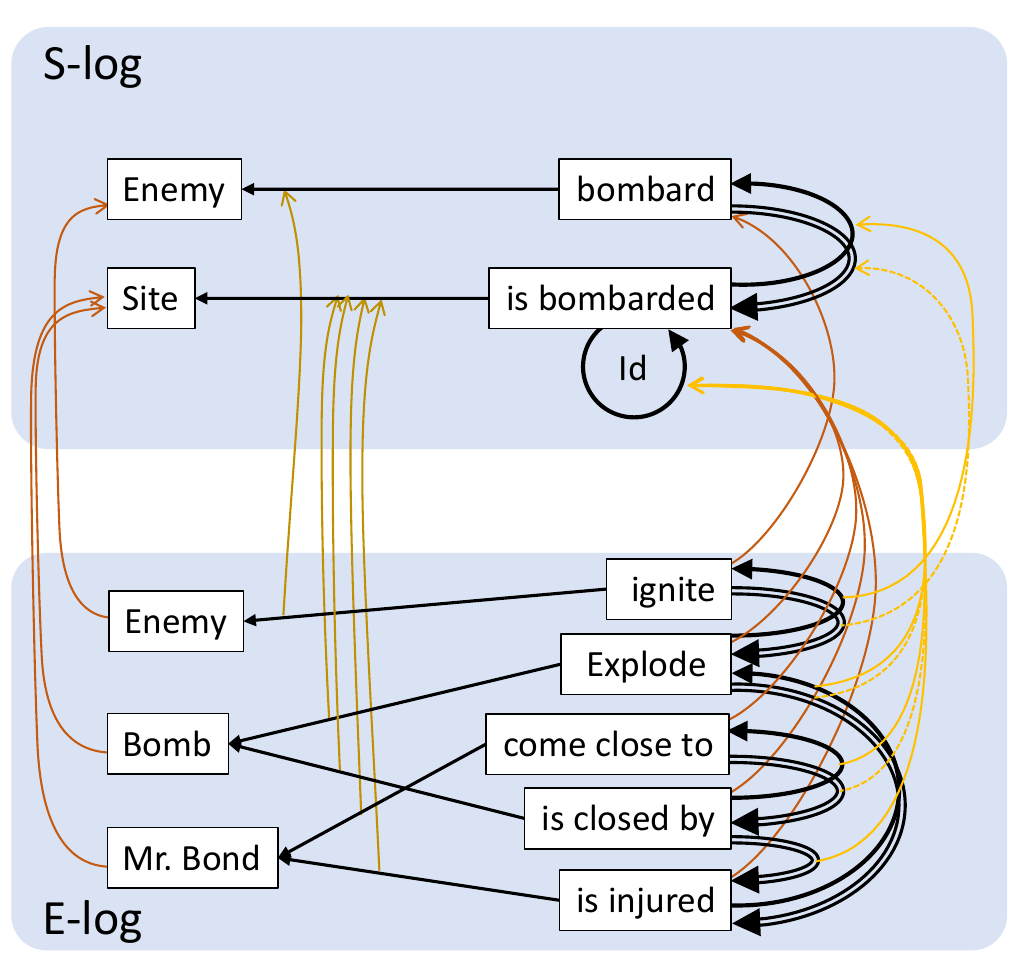}}% 
 \caption{Functors concerning a Mr. Bond's event. 
%injury according to the law 
%             that ``an explosion destroys things nearby" 
%             and abstractions o
             %a) Prediction of Mr. Bond's injury as filling in incomplete parts of the functor. 
                %from the e-log to the s-log
                %predicts occurrences in the near future. 
              Abstractions of the event into different scenarios. 
%             a) the functor from the e-log into the s-log omits the tower and the related actions. 
%                The actions ``\textit{destroy}" and ``\textit{is destroyed}" are not in the e-log. 
%             b) the actions ``\textit{destroy}" and ``\textit{is destroyed}" 
%                are copied to the e-log as future items. 
%                Then the functor is complete.  
    }
  \label{explosionfunctor2}
\end{figure}

Abstraction is one of the useful applications of functors among cognitive-logs. 
Consider a functor from an e-log which depicts a concrete event 
into an s-log which depicts an abstract scenario.  
Abstractions of the event into ``Enemy attack Mr. Bond" 
%``a robot carries a cart" into ``a worker carries a cargo" 
are illustrated in Fig. \ref{explosionfunctor}a. 
%In the abstraction shown in Fig. \ref{robotfunctor3}a, 
%the cart and its actions are mapped into the worker and its action ``carries," respectively. 
Note that some arrows 
%between these actions 
are mapped into the identity morphisms.  
Functors between cognitive-logs often exhibit arbitrariness. 
Figure \ref{explosionfunctor}b depicts another functor 
that maps into ``Enemy bombard the site." 
%that maps both the cart and the bottle into the cargo. 
%This functor implies that it does not matter what the robot carries. 
%While the functor in Fig. \ref{robotfunctor3}a focuses on the bottle, 
%and it does not matter what is used to carry it. 
The arbitrariness of abstraction models our ability to have multiple perspectives. 

Abstraction process can be repeated any number of times. 
It enables hierarchical abstraction. 
It should be noted that an inverse process of abstraction, 
abstract story into the details, is possible. 
The ``\textit{love}" in the story of Bob and Alice is described in this way. 

%A set of processes of abstraction and reconstruction represents a step of abstraction. 
%This step can be repeated any number of times. 
%An hierarchical abstraction is thus constructed. 
%Processes of opposite direction into details are also possible. 
%The ``\textit{love}" in the story of Bob and Alice is described in this way. 

\subsection{Planning and invention}

As a categorical operation, decomposition of an e-log into multiple e-logs is possible. 
And synthesis of multiple e-logs into one is also possible as the inverse of decomposition. 

Planning and invention can be modelled as a synthesis process of multiple s-logs. 
It consists of the following steps: 
%
%\begin{itemize}
%  \item 
Choose some s-logs from the storage of knowledge (s-logs), 
%  \item 
Assemble an s-log so that the scenario ends with a preferable result, 
%  \item 
and Convert the s-log into an e-log of the plan 
          by assigning each participant in the s-log to a participant (existing object) in the e-log. 
%\end{itemize}
%
Since the number of possible combinations of s-logs will be huge, 
the computational cost seems to be high. 
This is why such creative thinking is difficult.

%%%%%%%%%%%%%%%%%%%%%%%%%%%%%%%%%%%%%%%%%%%%%%%%%%%

\section{Functor search}

\subsection{Functor evaluation based on structure}

A functor is a structural matching between categories. 
Mathematical completeness is the basis of functor evaluation. 
A homomorphism in a category can be represented using a logical matrix 
whose entries and operations are Boolean algebra. 
Here, we define such matrices: $E$ as the arrows of ``\textit{Who,}" 
$S$ as the arrows of ``\textit{Cause,}" 
and $N$ as the arrows of ``\textit{Effect.}" 
The conversions of $S$ and $N$ in a functor from an e-log into an s-log satisfy the following equations:
\begin{equation}\label{S_new}%12
 \sum^\infty_{\acute{n}=1} S_s^{\acute{n}} + I_{Ss} 
   = P_S \left(\sum^\infty_{n=1} S_e^n \right) P_S^T + I_{Ss}.
\end{equation}
\begin{equation}\label{N_new}%13
 \sum^\infty_{\acute{n}=1} N_s^{\acute{n}} + I_{Ss} 
   = P_S \left(\sum^\infty_{n=1} N_e^n \right) P_S^T + I_{Ss}.
\end{equation}
Where, the suffix ``$_e$" denotes the e-log and ``$_s$" denotes the s-log respectively, 
$P_S$ is the conversion matrix that indicates the mappings between actions. 
The computation must use Boolean algebra; namely, $1+1=1$. 
$I_{Ss}$ is an identity matrix that represents the identity morphisms. 
Since these computations are Boolean, the sums with the identity matrix cannot be removed. 
The sums of the power series of $S_s$ and $S_e$ represent composites of ``\textit{cause}" arrows. 
Under the rule of causality, 
the ``\textit{cause}" and ``\textit{effect}"matrices $S$ and $N$ are strictly triangular matrix. 
Hence, $S_s^{\acute{n}}$ and $S_e^n$ become zero for finite $\acute{n}$ and $n$. 
The power series may rapidly decay.  

Here, arrows $S_{s,e}$ and $N_{s,e}$ can be divided into 
trivial parts---$S_{s,e}^{Tri}$ and $N_{s,e}^{Tri}$, 
and non-trivial parts---$S_{s,e}^{Non}$ and $N_{s,e}^{Non}$. 
Arrows of $m$-th pair of trivial causal relationship 
$S_{s,e\:m}^{Tri}$ and $N_{s,e\:m}^{Tri}$ are transposed matrices each other. 
An arrow of a pair of trivial causal relationship has only one non-zero entry. 
The trivial parts are regarded as sums of trivial causal relationships. 

Considering the exchange of ``\textit{cause}" and ``\textit{effect}" in trivial causal relationships, 
arrows $S_{s,e}$ and $N_{s,e}$ are replaced prior to the conversion as follows: 
\begin{equation}\label{S_exchange}%14
 S_{s,e} \leftarrow S_{s,e}^{Non} + \sum_{m=1} \left\{ r_m N_{s,e\:m}^{Tri} + (1-r_m) S_{s,e\:m}^{Tri} \right\}.
\end{equation}
\begin{equation}\label{N_exchange}%15
 N_{s,e} \leftarrow N_{s,e}^{Non} + \sum_{m=1} \left\{ r_m S_{s,e\:m}^{Tri} + (1-r_m) N_{s,e\:m}^{Tri} \right\}.
\end{equation}
where $r_m$ is the indicators that $r_m =1$ represents exchanging 
``\textit{cause}" and ``\textit{effect}" in $m$-th pair of trivial causal relationship. 

The conversion of $E$ in a functor from an e-log into an s-log is as follow:
\begin{equation}\label{E_new}%16
 E_s = P_E E_e P_S^T
\end{equation}
where $P_E$ is the conversion matrix that indicates the mappings between participants.  

Because the conversions are functions, 
there is one and only one entry of $P_E$ and $P_S$ in each column whose value is $1$, 
while the others are zero. 
Moreover, the converted e-log must satisfy its functional relationship from actions to participants. 
Namely, if a participant does not have its morphism function into the s-log, 
its action(s) cannot have a morphism function into the s-log. 
This requirement falls under the following mathematical rule: 
if $P_E E_e$ has a column with all entries being zero, 
then all entries in the corresponding column 
(the column number is the same as $P_E E_e$) of $P_S$ must be zero.

%S-logs are built so that all their objects are essential. 
%That means the functor must be surjective. % 全射
%A surjective functor satisfies the following requirement: 
%at least one entry in each row of $P_E$ and $P_S$ has to be non-zero. 

%If all objects in the e-log are essential, 
%the functor must be injective. % 単射
%An injective functor satisfies the following requirement: 
%at least one entry in each column of $P_E$ and $P_S$ has to be non-zero. 

A complete functor satisfies the above rules. 
Incompleteness of the functor may indicate incompatibility between the e-log and s-log, 
%perhaps the s-log was inadequate. 
%However, the incompleteness may indicate 
things that are going to happen, 
or hidden events that occurred in the past 
as discussed in \S \ref{inferences}. 
%An inference according to the laws (\S \ref{inferences}) applies to such a non-surjective functor. 

Finding functor is finding the conversion matrices $P_E$ and $P_S$ 
that satisfy the evaluations above. 
Search based on random generation and evaluation seem an appropriate way to find these matrices. 
Such a search requires computation cost, 
however, quantum computing is suitable for such searches, it might be a game changer. 
This is promising future item.

\subsection{Functor evaluation based on temporal order}\label{temporal_order}

\begin{figure}%8
\centering 
  \includegraphics[width=7.cm]{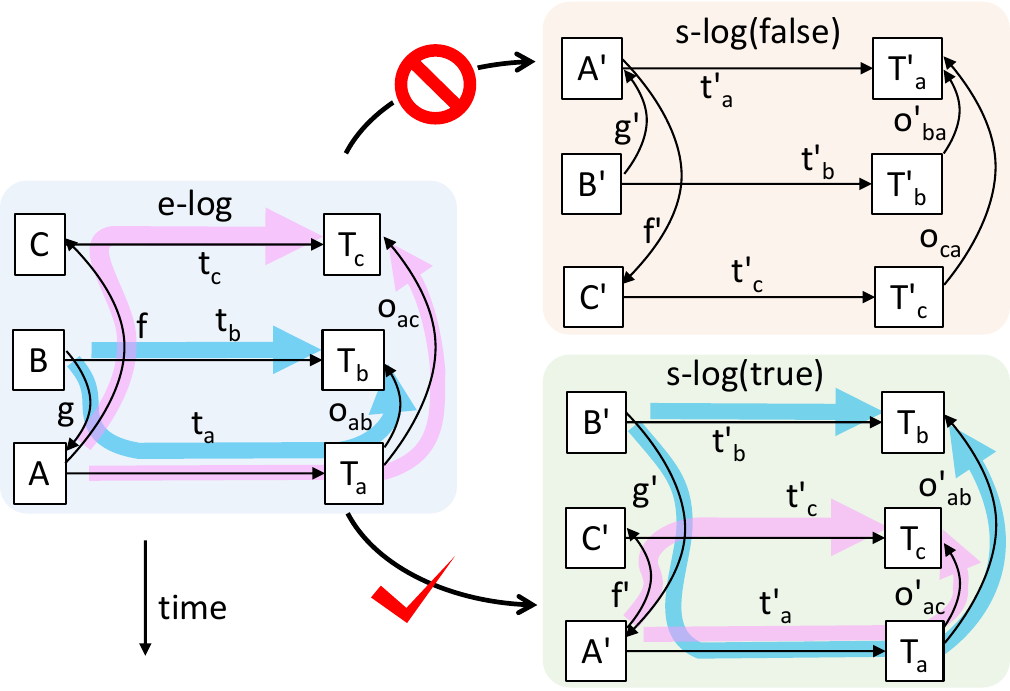}
 \caption{Functors from an e-log into s-logs. 
            A, B, and C represent actions, and $T_a$, $T_b$, and $T_c$ represent their timestamps. 
            This figure omits participants. 
%            In the e-log, composites of arrows and temporal orders are compatible 
%            in the functor into s-log (true). 
%            However, they are not compatible in the functor into s-log (false).  
 }
  \label{time_cause}
\end{figure}

Let assume that each action has timestamps. 
If an effect occurs before its cause, it violates the rule of causality. 
Comparing the order of the timestamps 
evaluates the consistency of a functor between an e-log and an s-log. 

Consider arrows from actions to their timestamps 
and assume there is an order relation between a pair of timestamps 
when there is a causal relationship between the corresponding actions. 
Figure \ref{time_cause} depicts an e-log and functors into s-logs. 
In this e-log, composites of arrows are commutative as follow: 
\begin{equation}\label{timeorder1}%17
 t_c \circ f = o_{ac} \circ t_a
\end{equation}
and
\begin{equation}\label{timeorder2}%18
 o_{ab} \circ t_a\circ g = t_b.
\end{equation}
Where $f$ and $g$ are the ``\textit{cause}" and ``\textit{effect}" arrows, respectively, 
$t_a$, $t_b$, and $t_c$ are arrows from actions to their timestamps, 
and $o_{ac}$ and $o_{ab}$ are arrows of the order relation between the timestamps 
that point from the future to the past. 
There is no causal relation between ``B" and ``C," 
hence the temporal order between them can be swapped. 
Conversely, the temporal order between ``A" and the others cannot be swapped. 
A consistent functor preserves the commutative relations 
in eqs. (\ref{timeorder1}) and (\ref{timeorder2}). 
The aforementioned functor evaluations based on eqs. (\ref{S_new}) and (\ref{N_new}) 
automatically satisfy consistency of temporal order. 

In realistic episodic memories, very few timestamps are recorded 
but limited information of temporal orders between actions are recorded. 
However, such temporal orders are important factors for logical reasoning. 
A set of temporal order information forms preorder, 
and a preorder is a category: t-logs. 
A t-log is an independent category, 
and it can be expressed as a sparse Boolean matrix $O_t$. 
If an action $i$ precede another action $j$ or they act in the same time, 
the $ij$-th entry is 1, i.e., $O_{tij} = 1$. 
A pair of trivial causal-relationship are identified as its corresponding entries are $1$, 
namely, $O_{tij} = O_{tji} = 1$. 
Evaluation can be done by comparing the transformed temporal order 
$P_S O_t P_S^T$ and the temporal order in the s-log $\acute{O}_t$.

\subsection{Functor evaluation based on similarity}

An event "Bob loves Alice" and another event "Mike hates Rob" have same structure. 
Since objects in cognitive-logs are recorded as token, 
the evaluations above provide only structural matching between cognitive-logs. 
In a functor between cognitive-logs, the pair of objects (domain and codomain) must be similar, 
i.e., evaluation of similarity between "loves" and "hates" is needed. 
%i.e., Mr. Bond must be similar to ``victim". 
%However, ``bomb" is not similar to ``enemy" nor ``site". 
In an abstraction, multiple objects are mapped into an object. 
For such a case, following rule seems appropriate: 
at least one object must be similar to the mapped object. 
Information in be-logs such as `actionA is "loves"' 
plays an important role for this evaluation. 
Its concrete algorithm is the future consideration.

%%%%%%%%%%%%%%%%%%%%%%%%%%%%%%%%%%%%%%%%%%%%%%%%%%%%%%%%%%%%
\section{Discussions}

\subsection{Logical reasoning}% and metaphor}

\begin{table*}
\caption{Logical reasoning and implementation using cognitive-logs}
\begin{center}
 \begin{tabularx}{160mm}{l|p{7em}|X|X} \hline
  \multicolumn{2}{l|}{Type} & Process & Process using cognitive-logs \\ \hline
  \multicolumn{2}{l|}{Deduction} 
   & Applying laws (premises) 
   & Finding functors between an e-log and an s-log, 
     or creation of a new s-log by composing existing s-logs \\ \hline
  \multirow{3}{*}{Induction} & Emumerative induction  
   & Regarding experience(s) as a general law 
   & Conversion of an e-log or e-logs into an s-log \\ \cline{2-4}
   & Abduction (retroduction) 
   & Creation of a hypothesis referring experience(s) 
   & Creation of an s-log, or filling an incomplete e-log or s-log \\ \cline{2-4}
  & Analogy \&  metaphor\cite{LakoffJohnson1980} 
  & Applying law(s) upon a case according to its similarity 
  & Finding a functor between an e-log and an s-log, 
     or creation of a new s-log by composing existing s-logs, 
     accepting low compatibility between corresponding objects \\ \hline
 \end{tabularx}\label{logictable}
\end{center}
\end{table*}

Operations using cognitive-logs represent logical inferences. 
The classification of logical inferences and operations using cognitive-logs 
is presented in table \ref{logictable}.

\subsubsection{Deductions}

Deduction, or deductive inference, is a type of logical inference process 
which draw conclusion from a premise. 
The aforementioned process of inferences according to laws is a type of deduction. 
An s-log depicts basic logics such as ``\textsf{if A then B}". 
%including mathematical theorems. 
And a functor search finds matchings. 
Composition of multiple s-logs and modification using functors 
is also a form of deduction that creates new theorems.

\subsubsection{Inductions}

Induction is a process of finding a law from experiences. 

Conversion of e-logs into an s-log may represents 
the most typical form of induction, known as enumerative induction. 
The most reasonable process involves gathering ``similar" experiences 
and storing their ``average" in an s-log. 
%Functor operations implement these similarity evaluations and averaging processes. 
If an experience was ``reasonable," even a single experience generates an s-log. 
Deductive inferences evaluate its ``reasonableness." 
Unlike reinforcement learning, 
humans do not require thousands of experiences. 
Generation of an s-log may resemble such human learning. 

Abduction or retroduction is also a kind of induction; 
It generates a hypothesis from experiences and other laws. 
This process may appear to fill in incomplete parts of an s-log or an e-log 
rather than generating a new one. 
This process may involve many deductions.

\subsubsection{Analogy and metaphor}

As Lakoff and Johnson demonstrated in their famous literature\cite{LakoffJohnson1980}, 
metaphors are 
%not only rhetoric in literary works but also 
important parts of cognitive processes. 
%According to them, most of the human conceptual system is made up of metaphors. 
%In the ``ARGUMENT IS WAR" example, 
%many elements in an argument are expressed using elements of war: 
%\textit{attack a position}, \textit{indefensible}, \textit{strategy}, 
%\textit{new line of attack}, \textit{win}, and \textit{gain ground}. 
%Metaphors ignore difference between objects, 
%the qualitative properties of war and that of argument, 
%such as the existence of violent factors of war: 
%death of humans, blood, pain, and others. 
%but preserves the logical structures. 
%of the concept of war, 
%i.e., the causal relationships between the elements of war. 
%
A metaphor ``maps" one event or concept onto another while 
preserving the structures of causal chains. 
Such a metaphor systematically ignores the similarities between objects. 
%some of these properties. 
%Functors are structure preserving maps. 
A functor between cognitive-logs with modifications in similarity evaluation may resemble a metaphor.

\subsection{Feedback to cognitive sciences}

\subsubsection{Volition and personification}

As shown in Fig. \ref{schema}, volition (intention) always exists prior to a human's action. 
%When people see a robot carries a cart, they may imagine the volition of its operator.  
Unlike ``cube $A$ is on top of cube $B$", 
the actual relationship between Bob and Alice in the event ``Bob loves Alice" is not symmetric 
because there is Bob's volition acting on Alice. 
It should be noted that quite a lot of verbs assume existence of a volition. 

We often use personifications for understanding physical process. 
%As, we often use the word ``behaviour" for expressing physical properties of an object. 
It is difficult to remove personification from our understanding of physics. 
%Personification is useful 
%because it allows application of \textit{theory of mind} to the event. 
%Theory of mind is inherent ability to infer other's mental states. 
%Personification for an object gives consistency to the behaviour of the object 
%and make the behaviour predictable. 

%As mentioned before, 
We discretize the world. 
Surface of objects define the boundaries in space, however, 
action boundaries over time are hard to define.  
%We have a world-model that actions are discrete. 
%This is because we have discrete volitions. 
%Although each motion in the real-world is continuous, 
%we switch our own volitions (tasks). 
Perhaps, personification is a part of our inherent world-model, 
and discrete switch of volition define action boundaries. 
%that the model---``actions are discrete"---is applicable to any object in the world. 
%Logical reasoning is based on the cognition of causal chains of actions. 
%Cognition of actions is essential, 
%and personification---assuming volition---helps in this regard. 
\cite{tomasello1999} presents a hypothesis that 
the development of the cognition of others' volition initiated 
the development of language. 
The basis of understanding causality in infants is the assumption of volition 
even for non-living things. 
If cognition of volition and personification are essential part of intelligence, 
are they essential for artificial intelligence too? 

Incorporation of those social cognition into an artificial intelligence is apparently complex and unrealistic. 
However, it is also questionable to establish an intelligence with omitting such essential cognition. 
It is for future consideration. 

\subsubsection{Understanding}

The film ``Tron" released in 1982, 
%and famous for being one of the first to adopt computer graphics, 
depicts a battle within a computer system which is advanced concepts in that era. 
%The film introduced many advanced %and innovative 
%concepts for computers. %and networks. 
Unfortunately, the film was not successful. 
%in the business. 
Audiences in the 1980s did not feel a sense of comprehension for the story. 
%Such such advanced concepts and scenario were not well-known in that era. 

%
Understanding of an situation is modelled as searching for a functor from the e-log into an s-log, 
where the s-log represents a familiar scenario.  
%The feeling of ``understanding" is modelled as obtaining high completeness of the functor. 
According to a research\cite{Kurashige2018}, 
our acquisition of knowledge depends on prior experience 
and correlations to prior experience play an important role. 
%The process of finding a functor from e-logs into s-logs 
%may be useful tool for modelling such a cognitive process. 
Cognitive-log may model such processes.

\subsubsection{Two systems}

 ``When you have eliminated the impossible, whatever remains, 
however improbable, must be the truth."%
---Arthur Conan Doyle, The Case-Book of Sherlock Holmes. 
This short quote shows that 
rigorous logic sometimes appears to be improbable in our intuitive reasoning. It 
may indicate that neural-networks are inefficient for logical reasoning. 

\cite{Kahneman2011} presented the idea 
that our thinking operates through two distinct systems. 
System 1 is fast and intuitive, 
while System 2 is slow but accurate and logical. 
Neural-network-based artificial intelligences resemble System 1, 
and reasoning based on cognitive-logs may resemble System 2.

A trained LLM with operations on cognitive-logs 
may resemble logical reasoning under System 1. 
On the other hand, operations using category theory on cognitive-logs 
can provide rigorously correct answers, much like Sherlock Holmes.

\section{Concluding remarks}

Neural-networks are not efficient for logical reasoning. 
This study presents a new framework for describing episodic memories and logical reasoning. 
It was shown that 
cognitive-logs and the processes within them, especially functor searches, 
provide mathematical models for various reasoning. 
There is a possibility that these operations offer models of human mind.

Implementation of these operations in a computer system consists of relational databases 
and search for conversion matrices. 
It enables a database-driven artificial intelligence that thinks like a human 
but possesses the accuracy and rigour of a machine. 
Quantum computing seem suitable for the functor search operations within cognitive-logs. 
The vast capacity of database and the enormous power of quantum computing promise the potential of cognitive-logs. 

The findings from cognitive linguistics clarified that actions are the main component of episode recognition. 
Thus, action is the primary key.

\section*{Ethical Statement}

There are no ethical issues.

\section*{Acknowledgments}
This research received no specific grant from any funding agency 
in the public, commercial, or not-for-profit sectors.

The author thanks Hideki Kajima and people in Frontier Research Centre in Toyota Motor Corporation 
for their assistance to the author's study.

\bibliographystyle{named}
\bibliography{ijcai25_fukada}

\begin{thebibliography}{}

\bibitem[\protect\citeauthoryear{Bain}{2016}]{Bain2016}
R.~Bain.
\newblock Are our brains bayesian?
\newblock {\em Significance}, 13(4):14--19, 2016.

\bibitem[\protect\citeauthoryear{Bakker}{1987}]{Bakker1987}
R.R. Bakker.
\newblock {\em Knowledge Graphs: Representation and Structuring of Scientific
  Knowledge}.
\newblock PhD thesis, University of Twente, 1987.

\bibitem[\protect\citeauthoryear{Bronson \bgroup \em et al.\egroup
  }{2013}]{TAO}
Nathan Bronson, Zach Amsden, George Cabrera, Prasad Chakka, Peter Dimov~Hui
  Ding, Jack Ferris, Anthony Giardullo, Sachin Kulkarni, Harry Li, Mark
  Marchukov~Dmitri Petrov, Lovro Puzar, Yee~Jiun Song, and Venkat
  Venkataramani.
\newblock Tao: Facebook’s distributed data store for the social graph.
\newblock pages 49--60, 2013.

\bibitem[\protect\citeauthoryear{Codd}{1970}]{Codd1970}
E.~F. Codd.
\newblock A relational model of data for large shared data banks.
\newblock {\em Communications of the ACM}, 13(6):377--387, 1970.

\bibitem[\protect\citeauthoryear{Croft}{1991}]{croft1991}
W.~Croft.
\newblock {\em Syntactic Categories and Grammatical Relations: The Cognitive
  Organization of Information}.
\newblock University of Chicago Press, Chicago, 1991.

\bibitem[\protect\citeauthoryear{Feigenbaum}{1984}]{Feigenbaum1984}
E.A. Feigenbaum.
\newblock Knowledge engineering: The applied side of artificial intelligence.
\newblock {\em Annals of the New York Academy of Sciences}, 426(1):91--107,
  1984.

\bibitem[\protect\citeauthoryear{Fuyama \bgroup \em et al.\egroup
  }{2020}]{Fuyama2020}
M.~Fuyama, H.~Saigo, and T.~Takahashi.
\newblock A category theoretic approach to metaphor comprehension: Theory of
  indeterminate natural transformation.
\newblock {\em Biosystems}, 197:104213, 2020.

\bibitem[\protect\citeauthoryear{Hopper and
  Thompson}{1980}]{hooperThompson1980}
P.~Hopper and S.~Thompson.
\newblock Transitivity in grammar and discourse.
\newblock {\em Language}, 56(2):251--299, 1980.

\bibitem[\protect\citeauthoryear{Kahneman}{2011}]{Kahneman2011}
D.~Kahneman.
\newblock {\em Thinking, fast and slow}.
\newblock Farrar, Straus and Giroux, New York, 2011.

\bibitem[\protect\citeauthoryear{Kowalski}{1988}]{Kowalski1988}
R.A. Kowalski.
\newblock The early years of logic programming.
\newblock {\em Communications of the ACM}, 31(1):38--43, 1988.

\bibitem[\protect\citeauthoryear{Kurashige \bgroup \em et al.\egroup
  }{2018}]{Kurashige2018}
Hiroki Kurashige, Yuichi Yamashita, Takashi Hanakawa, and Manabu Honda.
\newblock A knowledge-based arrangement of prototypical neural representation
  prior to experience contributes to selectivity in upcoming knowledge
  acquisition.
\newblock {\em Frontiers in Human Neuroscience}, 12, 2018.

\bibitem[\protect\citeauthoryear{Ladkin and Loer}{1998}]{ladkin1998}
P.~Ladkin and K.~Loer.
\newblock Analysing aviation accidents using wb-analysis - an application of
  multimodal reasoning, 1998.

\bibitem[\protect\citeauthoryear{Lakoff and Johnson}{1980}]{LakoffJohnson1980}
G.~Lakoff and M.~Johnson.
\newblock {\em Metaphors we live by}.
\newblock U. of Chicago Press, Chicago, 1980.

\bibitem[\protect\citeauthoryear{Lakoff}{1987}]{Lakoff1987}
G.~Lakoff.
\newblock {\em Women, fire and dangerous things: what categories reveal about
  minds}.
\newblock University of Chicago Press, Chicago, 1987.

\bibitem[\protect\citeauthoryear{Langacker}{1987}]{Langacker1987}
R.~W. Langacker.
\newblock Nouns and verbs.
\newblock {\em Language}, 63(1):53--94, 1987.

\bibitem[\protect\citeauthoryear{Le\:Cun}{2022}]{LeCun2022}
Yann Le\:Cun.
\newblock A path towards autonomous machine intelligence version 0.9.2,
  2022-06-27, 2022.

\bibitem[\protect\citeauthoryear{Mirzadeh \bgroup \em et al.\egroup
  }{2024}]{Mizadeh2024}
Iman Mirzadeh, Keivan Alizadeh, Hooman Shahrokhi, Oncel Tuzel, Samy Bengio, and
  Mehrdad Farajtabar.
\newblock Gsm-symbolic: Understanding the limitations of mathematical reasoning
  in large language models, 2024.

\bibitem[\protect\citeauthoryear{Ohori}{2002}]{Ohori}
T.~Ohori.
\newblock {\em Cognitive Linguistics (in Japanese)}.
\newblock University of Tokyo Press, Tokyo, 2002.

\bibitem[\protect\citeauthoryear{Spivak and Kent}{2012}]{Spivak2012}
D.I. Spivak and R.E. Kent.
\newblock Ologs: A categorical framework for knowledge representation.
\newblock {\em PLoS ONE}, 7(1):e24274, 2012.

\bibitem[\protect\citeauthoryear{Spivak}{2010}]{databaseiscategory}
D.~I. Spivak.
\newblock Databases are categories, 2010.

\bibitem[\protect\citeauthoryear{Spivak}{2012}]{databaseiscategory2}
D.~I. Spivak.
\newblock Functorial data migration.
\newblock {\em Information and Computation}, 217:31--51, 2012.

\bibitem[\protect\citeauthoryear{Spivak}{2014}]{spivak_text}
D.~I. Spivak.
\newblock {\em Category Theory for the sciences}.
\newblock MIT Press, Cambridge, Massachusetts, 2014.

\bibitem[\protect\citeauthoryear{Tomasello}{1999}]{tomasello1999}
M.~Tomasello.
\newblock {\em The cultural origins of human cognition}.
\newblock Harvard University Press, Harvard, 1999.

\bibitem[\protect\citeauthoryear{Tomasello}{2000}]{Tomasello2000}
M.~Tomasello.
\newblock Do young children have adult syntactic competence?
\newblock {\em Cognition}, 74(3):209--253, 2000.

\bibitem[\protect\citeauthoryear{Tversky}{1977}]{Tversky1977}
A.~Tversky.
\newblock Features of similarity.
\newblock {\em Psychological review}, 84:327--352, 1977.

\bibitem[\protect\citeauthoryear{Vaswani \bgroup \em et al.\egroup
  }{2017}]{vaswani2017attention}
Ashish Vaswani, Noam Shazeer, Niki Parmar, Jakob Uszkoreit, Llion Jones,
  Aidan~N Gomez, \L~ukasz Kaiser, and Illia Polosukhin.
\newblock Attention is all you need.
\newblock In I.~Guyon, U.~Von Luxburg, S.~Bengio, H.~Wallach, R.~Fergus,
  S.~Vishwanathan, and R.~Garnett, editors, {\em Advances in Neural Information
  Processing Systems}, volume~30. Curran Associates, Inc., 2017.

\bibitem[\protect\citeauthoryear{Wang \bgroup \em et al.\egroup
  }{2024}]{Wang2024}
Kevin Wang, Junbo Li, Neel~P. Bhatt, Yihan Xi, Qiang Liu, Ufuk Topcu, and
  Zhangyang Wang.
\newblock On the planning abilities of openai's o1 models: Feasibility,
  optimality, and generalizability, 2024.

\bibitem[\protect\citeauthoryear{Woods}{1975}]{Woods1975}
W.A. Woods.
\newblock {\em What's in a Link: Foundations for Semantic Networks in
  Representation and Understanding}.
\newblock Morgan Kaufmann, San-Diego, CA, 1975.

\bibitem[\protect\citeauthoryear{Wynn}{1992}]{Wynn1992}
K.~Wynn.
\newblock Addition and subtraction by human infants.
\newblock {\em Nature}, 358:749--750, 1992.

\bibitem[\protect\citeauthoryear{Yamakawa}{2021}]{Yamakawa_2021}
Hiroshi Yamakawa.
\newblock The whole brain architecture approach: Accelerating the development
  of artificial general intelligence by referring to the brain.
\newblock {\em Neural Networks}, 144:478–495, December 2021.

\end{thebibliography}

\end{document}